\relax
\documentclass[letterpaper]{article} 
\usepackage{aaai22}  
\usepackage{times}  
\usepackage{helvet}  
\usepackage{courier}  
\usepackage[hyphens]{url}  
\usepackage{graphicx} 
\usepackage{natbib}  
\usepackage{caption} 
\DeclareCaptionStyle{ruled}{labelfont=normalfont,labelsep=colon,strut=off} 
\usepackage{bm}
\usepackage{amsmath}
\usepackage{amssymb}
\usepackage{multirow}

\usepackage{algorithm}
\usepackage{algorithmic}

\makeatletter
\newcommand{\thickhline}{%
    \noalign {\ifnum 0=`}\fi \hrule height 1pt
    \futurelet \reserved@a \@xhline
}
\makeatother

\usepackage{newfloat}
\usepackage{listings}
\floatstyle{ruled}
\newfloat{listing}{tb}{lst}{}
\floatname{listing}{Listing}

\title{End-to-End Segmentation-based News Summarization}
\author{
    Yang Liu,
    Chenguang Zhu,
    Michael Zeng
}
\affiliations{
    Microsoft Cognitive Services Research Group\\
    \{yaliu10,  chezhu, nzeng\}@microsoft.com
}

\usepackage{bibentry}

\begin{document}

\maketitle

\begin{abstract}
In this paper, we bring a new way of digesting news content by introducing the task of segmenting a news article into multiple sections and generating the corresponding summary to each section. 
We make two contributions towards this new task. First, we create and make available a dataset, \textsc{SegNews}, consisting of 27k news articles with sections and aligned heading-style section summaries. Second, we propose a novel segmentation-based language generation model adapted from pre-trained language models that can jointly segment a document and produce the summary for each section. 
Experimental results on \textsc{SegNews} demonstrate that our model can  outperform several state-of-the-art sequence-to-sequence generation models for this new task.
\end{abstract}

\section{Introduction}
In recent years, automatic summarization has received extensive attention in the
natural language processing community, due to its
potential for processing redundant information.
The evolution of neural network models and availability of large-scale
datasets have driven the rapid development of summarization systems.

Despite promising results, 
there are specific characteristics of the traditional
summarization task that impedes it to provide more beneficial ways of digesting long news articles.  
For instance, current news summarization system only provides one genetic summary of the whole article, and when users want to read in more details, the generated summary is not capable of helping navigate the reading.
For example, given a  news report, current system will output several highlight summaries~\cite{nallapati2017summarunner, liu2019text, zhang2020pegasus}.
Under this circumstance, if a user expect to read more details about one highlight, he will still need to browse the whole article to locate related paragraphs.  
Meanwhile, when processing a long news article, current systems usually truncate the text and only generate a summary based on the partial article~\citep{cheng-lapata:2016:P16-1, zhang2020pegasus}. Although this is reasonable since most important content usually lies in the initial portion, it also makes it difficult for users to quickly access information beyond the truncated portion.

In this paper, we propose a new task of  Segmentation-based News Summarization.
Given a news article, we aim to identify its
potential sections and at the same time, to generate the corresponding summary for each section.
This new task provides a novel alternative to summarizing a news article. We argue that it can lead to a more organized way of understanding long articles and facilitates a more effective style of reading documents. 

\begin{figure*}[t]
    \begin{center}
    \includegraphics[width=1\textwidth]{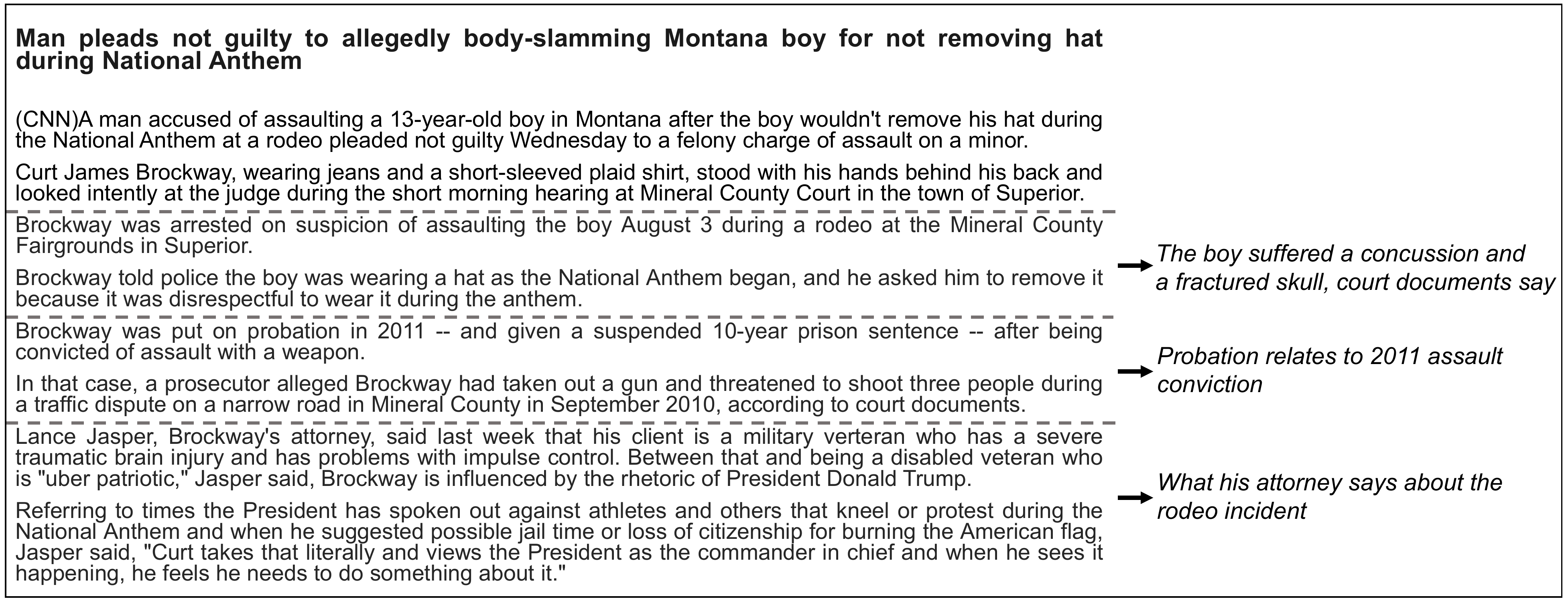}
    \caption{One example from the segmentation-based summarization task \textsc{SegNews}. The news article is taken from a CNN news article and we truncate the article for display. CNN editors have divided this article into several sections and written a heading to section. The goal of this task is to automatically identify sub-topic segments of multiple paragraphs, and  generate the heading-style summary for each segment. Dotted lines in the figure indicate segment boundaries. In this article, paragraphs 1,2 are annotated as the first segment,  paragraphs 3,4 are annotated as the second segment, paragraphs 5,6 are annotated as the third segment, and paragraphs 7,8 are annotated as the forth segment. To the right of the article are the  heading-style summaries for segments. Since the first segment is usually an overview  of the news, we do not assign a summary to it.\label{fig1}}
    \end{center}
\end{figure*}

First, segmenting a news article can provide a structural organisation of the content, which is not only helpful to reading but also benefit many important NLP tasks.
For example, 
~\citet{brown1983discourse} states that this kind of multi-paragraph division is one of the most fundamental tasks in discourse. 
However, many expository texts, like news articles, instruction manuals, or textbooks consist of long sequences of paragraphs with very little structural demarcation~\cite{hearst-1994-multi}, and for these documents a subtopical segmentation can be useful. 
Second, generating concise text descriptions of each sections further
reduces the cognitive burden of reading the article~\cite{florax2010contributes}. Previous studies~\cite{paice1990constructing,hearst1997text} present that subtopic segments with their headings is an effective alternative to traditional summarization tasks.

In this paper, we make two main contributions towards the development of Segmentation-based News Summarization systems.

First, we create and publicize a large-scale benchmark\footnote{We will share the dataset link after paper acceptance.}, \textsc{SegNews}, for Segmentation-based News Summarization task. 
Figure~\ref{fig1} shows one example article and its aligned segmentation and summaries from \textsc{SegNews}.

Second, we propose a novel end-to-end approach for this task, which can jointly segment an article while generating the corresponding summaries.
These two sub-tasks can learn from each other via a shared encoder.
The model is equipped with a segmentation-aware attention mechanism, allowing it to capture segmentation information during summary generation.
One important advantage of our framework is that it is a non-invasive adaptation of the Transformer~\cite{vaswani2017attention} model, i.e. it does not alter the inner structure of Transformers. 
And our framework can integrate many pretrained language generation models, including \textsc{Bart}~\cite{lewis2019Bart}, \textsc{GPT}~\cite{radford2019language} and \textsc{UniLM}~\cite{bao2020UniLMv2}. This enables our framework to enjoy a high degree of flexibility and better performance.

We compare the proposed framework with several state-of-the-art methods on the \textsc{SegNews} benchmark.
Both automatic evaluation and human evaluation demonstrate the superiority of our model.

\section{Related Work}
\subsection{Document Summarization}

Document summarization is the task of automatically generating a shorter version text of one or multiple documents while retaining its most important information~\cite{radev2002introduction}. The task has received much attention in the
natural language processing community due to its
potential for various information access applications. 
Most large-scale summarization datasets are built on news articles. Popular single-document summarization benchmarks include CNN/DM~\cite{nallapati2016abstractive,cheng-lapata:2016:P16-1}, NYT~\cite{durrett2016learning} and XSum~\cite{xsum}.

Document summarization  can be classified into different paradigms by different factors~\cite{Nenkova:McKeown:2011}.
And among them,  two have consistently attracted attention.  
\textit{extractive} approaches form summaries by
copying and concatenating the most important spans in a document; while in \textit{abstractive}
summarization, various text rewriting operations generate
summaries using words or phrases that are not in the original
text. 

Recent approaches to
extractive summarization frame the task as a sequence labeling
problem by taking advantage of the success of neural network
architectures \cite{bahdanau:2015:iclr}. The idea is to predict a
label for each sentence specifying whether it should be included
in the summary.  Existing systems mostly rely on recurrent neural
networks~\cite{hochreiter:1997:nc} or Transformer model~\cite{vaswani2017attention} to encode the document and
obtain a vector representation for each
sentence~\citep{nallapati2017summarunner,cheng2016neural,yang19sumo}.

In recent years, neural sequence-to-sequence approaches dominate abstractive summarization methods.
\citet{rush2015neural} and \citet{nallapati2016abstractive} are among
the first to apply the neural encoder-decoder architecture to text
summarization.  \citet{see-acl17} enhance this model with a
pointer-generator network and a coverage mechanism. 
     Pretrained language models have recently emerged as a key
     technology for improving abstractive
     summarization systems.
     These models first pretrain a language model with self-supervised
     objectives on large corpora and then fine-tune it on
     summarization datasets.  \citet{liu2019text} combine a pretrained
     encoder based on  \textsc{Bert} \cite{devlin2018bert} with a randomly
     initialized decoder, demonstrating substantial gains on
     summarization performance.  \textsc{Mass}~\cite{song2019mass} is an
     encoder-decoder neural model pretrained with the objective of reconstructing a masked text and can be fine-tuned on summarization tasks.
     \textsc{Bart}~\cite{lewis2019Bart} is an encoder-decoder Transformer~\cite{vaswani2017attention} pretrained by reconstructing a text corrupted with several
     arbitrary noising functions.
 \citet{bao2020UniLMv2} design
     \textsc{UniLM}v2, a Transformer-based neural network pretrained
     as a pseudo-masked language model. 

\subsection{Text Segmentation and Outline Generation}

Text segmentation has been widely used in the fields of natural language processing and information extraction. 
Existing methods for text segmentation fall into two categories: unsupervised and supervised.
TextTiling~\cite{hearst1997text} is one of the first unsupervised topic segmentation algorithms.
It segments texts in linear time by calculating the similarity 
between two blocks of words based on the cosine similarity.
\citet{choi2000advances} introduce a statistical model which can calculate the maximum-probability segmentation of a given text.
The TopicTiling~\cite{riedl2012topictiling}  algorithm is based on  TextTiling, which uses the Latent Dirichlet 
Allocation to find topical changes
within documents.
LCSeg~\cite{galley2003discourse} computes lexical chains of documents and segments texts by a score which captures the sharpness of the change in lexical cohesion.

Supervised methods have also been proposed for text
segmentation. \citet{hsueh2006automatic} integrate lexical and
conversation-based features for topic and sub-topic segmentation. \citet{hernault2010sequential} use CRF to train a discourse
segmenter with a set of lexical and syntactic features. \citet{li2018segbot} propose \textsc{SegBot} which uses a neural network model with a bidirectional recurrent neural
network  together with a pointer network to select text boundaries in the input sequence.

Recently, \citet{zhang2019outline} propose Outline Generation task, aiming to identify  potential sections of a multi-paragraph document and generate the corresponding section headings as outlines. 
This task is in form similar to segmentation-based summarization. However, there are two main differences. First, outline generation focused on academic or encyclopaedic documents, where the section headings are extremely short (on average less than two words) and cannot be considered as a summarization task. Second, since outlines care more about briefly describing their  corresponding sections,  headings in  outlines are  independently from each other. In segmentation-based summarization task, despite describing the sections, heading-style summaries also devote to navigating the reading, and they are usually related and coherent in content.

\begin{table}[]
\renewcommand\arraystretch{1.05}

\begin{center}
\begin{tabular}{l|c}
\thickhline
\# news articles & 26,876\\
\#  paragraphs & 40.31\\
\#  sections per article & 3.17\\
\#  tokens per article & 1362.24\\
\#  tokens per section summary & 4.70\\
\thickhline
\end{tabular}
\end{center}

    \caption{Data statistics of the \textsc{SegNews} dataset. }
\end{table}

\section{The \textsc{SegNews} Benchmark}
In order to study and evaluate the Segmentation-based News Summarization task, we build a new benchmark dataset \textsc{SegNews}. 
We take CNN website as our article source.
As illustrated in Figure 1, there are a large part of CNN news articles which are divided by human editors into several sub-topic sections~(see Appendix for details).
And each section is assigned a heading-style summary also written by these editors.
We collect news articles published from 2017 to 2021, covering multiple CNN news channels, including US Politics, World, Business, Health, Entertainment, Travel and Sports.
We filter articles with no sub-topic structures or editor written heading-style summaries.
Since the first segment is usually an overview of the news, editors do not assign a summary to it.
The resulting dataset contains 26,876 news articles. For each article, it has human annotated segmentation structures and each segment has a human-written heading-style summary.

Table 1 shows the overall statistics of our \textsc{SegNews}
benchmark dataset. 
We can see that the news articles  in  \textsc{SegNews} contain rich structural information and are much longer (1,362 tokens per article) than traditional news summarization datasets: articles in CNN/DM~\cite{cheng2016neural} dataset has an average length of 686.63 tokens and articles in NYT~\cite{nytcorpus} dataset has an average length of 800.04 tokens. This is in line with our motivation that segmentation-based summarization can help readers better understand longer articles.


\begin{figure*}[htbp]
    \begin{center}
    \includegraphics[width=0.75\textwidth]{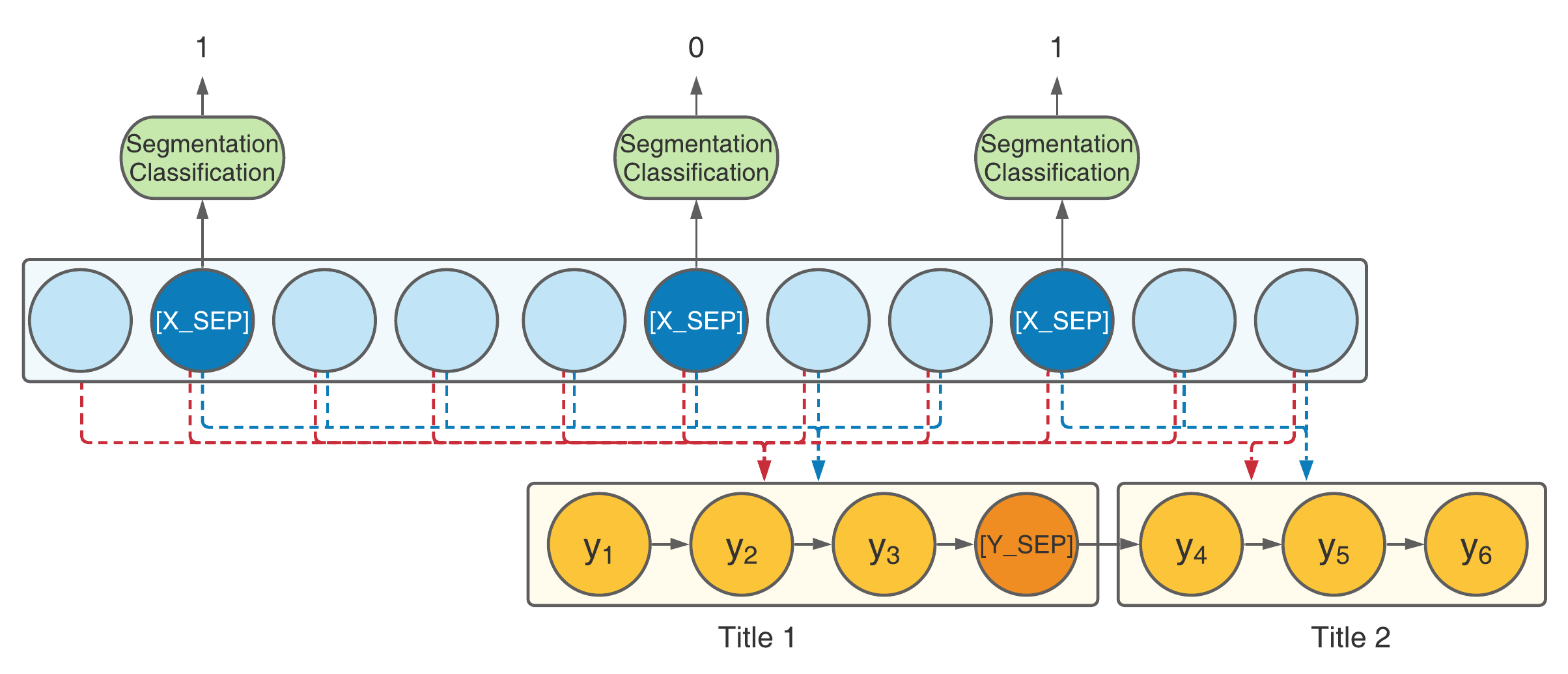}
    \caption{The overall framework of \textsc{SegTrans} model. The blue circles indicate input source text, where dark blue circles indicate special symbols representing paragraph boundaries. The yellow circles indicate output target text, where orange circles indicate special symbols representing title boundaries. Dotted blue lines indicate attention heads with segmentation-aware attention mechanism and red dotted lines indicate attention heads with original full attention mechanism. \label{fig2}}
    \end{center}
\end{figure*}

\section{Task Formulation }
Given a multi-paragraph article, the segmentation-based summarization task aims to: i)
identify sections of the article to 
unveil its
inherent sub-topic structure, where each section consists of neighboring paragraphs with a coherent topic, and ii) generate the heading-style summary for each section to concisely
summarize the section. 
Particularly, in one article, 
summaries of different sections should be coherent in content and consistent in style.

Formally, let $d$ indicate a document consisting of  paragraphs $[p_1, p_2, . . . , p_M]$.
The segmentation-based summarization task aims to recognize a sequence of section boundaries $[b_1, b_2, \cdots, b_{N-1}]$. These  boundaries divide the document into $N$ sections $s_1=[p_1, ..., p_{b_1}], s_2=[p_{b_1+1}, ..., p_{b_2}],\cdots, s_N=[p_{b_{N-1}+1}, ..., p_M]$.
Meanwhile, summarization systems will  generate the corresponding section summaries $[y_1, y_2, . . . , y_N]$.


\section{Systems for Segmentation-based News Summarization }
In this section, we present two different frameworks to tackle the segmentation-based summarization task. In the \textit{Pipeline} approach, we  first apply a segmentation 
model to identify
the potential sections, and then apply a generation model to produce the headings.
In the \textit{Joint}  approach, one neural model is able to jointly segment an article and produce the summaries.
To achieve this, we  design a novel segmentation-aware attention mechanism, which allows the model to capture segmentation information when generating summaries.
This new attention mechanism can also be considered as  a non-invasive adaption for  conventional Transformer models. Thus, to take the most advantage of existing pre-trained language models, we propose \textsc{SegUniLM} and  \textsc{SegBart} which are respectively based on pre-trained \textsc{UniLM} model and \textsc{Bart} model. They can be initialized completely from  pre-trained models and achieve substantial improvement on segmentation-based summarization task. 

\subsection{Pipeline Approach}
\paragraph{Segmentation model}
We formulate the section identification process as a sequence labeling task. We insert a special symbol \texttt{[X\_SEP]} at the boundary of paragraph $p_i$ and $p_{i+1}$ , and then concatenate all paragraphs into a single text input. A neural encoder is then applied to encode this input. 
Define $u_i$ as the output vector of \texttt{[X\_SEP]} after paragraph $p_i$. We then apply a binary classification layer over $u_i$ to obtain $y_i \in \{0, 1\}$. $y_i=0$ indicates paragraph $p_i$ and $p_{i+1}$ are in one segmentation,  and $y_i=1$ indicates $p_{i+1}$ should be the start of a new segment.

\paragraph{Generation model} We then generate an aligned heading-style summary for each identified section $s_j$. The generation of each heading is independent. Here, we can choose existing extractive or abstractive summarization methods.
\begin{itemize}
    \item \textsc{TopicRank}~\cite{bougouin2013topicrank} is an extractive method for keyphrase extraction which represents a document as a complete graph depending on topical representations. We use the top ranked phrase as the summary for input section;
    \item \textsc{Seq2Seq} represents the sequence-to-sequence neural model, which is usually used in abstractive summarization. It  first encodes the concatenated text of all paragraphs within this section, and the decodes the heading in an auto-regressive manner. In experiments, we try both non-pretrained Transformer model and pretrained \textsc{UniLM} and \textsc{Bart} models as \textsc{Seq2Seq} models.

\end{itemize}

\begin{table*}[]
\renewcommand\arraystretch{1}
\centering
\begin{tabular}{c|c|cccccc}
\thickhline
\multicolumn{2}{c|}{\textit{Vanilla Seq2Seq}}           & \multicolumn{2}{c}{R1}                      & \multicolumn{2}{c}{R2}                      & \multicolumn{2}{c}{RL}                       \\ \hline
\multicolumn{2}{c|}{\textsc{Trans}}                            & \multicolumn{2}{c}{8.66}                        & \multicolumn{2}{c}{1.51}                        & \multicolumn{2}{c}{8.16}                         \\
\multicolumn{2}{c|}{\textsc{Unilm} }                      & \multicolumn{2}{c}{19.22}                        & \multicolumn{2}{c}{7.18}                        & \multicolumn{2}{c}{16.99 }                         \\ \hline
\multicolumn{2}{c|}{\multirow{2}{*}{\textit{Pipeline}}} & \multicolumn{3}{c}{\multirow{2}{*}{With Gold Segments}}                  & \multicolumn{3}{|c}{\multirow{2}{*}{With Predicted Segments}} \\
\multicolumn{2}{c|}{}                          & \multicolumn{3}{c}{}                                                     & \multicolumn{3}{|c}{}                                         \\ \hline
Segmentor              & Generator             & R1                   & R2                   & \multicolumn{1}{c|}{RL}    & R1             & R2                    & RL                   \\ \hline

Transformer     & Transformer    & \multicolumn{1}{c}{8.69} & \multicolumn{1}{c}{1.83} &    \multicolumn{1}{c|}{9.09}          &   --               &    --     &--               \\
Transformer     & TopicRank    & \multicolumn{1}{c}{5.09} & \multicolumn{1}{c}{1.14} &    \multicolumn{1}{c|}{6.28}          &        --          &   --     &  --          \\
\textsc{Bart}           & \textsc{Bart}          & 21.42                & 7.76                 &  \multicolumn{1}{c|}{19.28} &     16.01     &   5.27  & 14.37               \\ 
\textsc{UniLM}           & \textsc{UniLM}          &21.76               & 8.22                & \multicolumn{1}{c|}{19.75} &    16.27    &      5.45  &  14.65              \\ 

\hline

\multicolumn{2}{c|}{\textit{Joint}}                     & \multicolumn{1}{c}{R1}                   & \multicolumn{1}{c}{R2}                   & \multicolumn{1}{c|}{RL}    & \multicolumn{1}{c}{R1}                   & \multicolumn{1}{c}{R2}                   & \multicolumn{1}{c}{RL}                     \\ \hline
\multicolumn{2}{c|}{\textsc{SegTrans}}      &  8.94          &     1.85                               & \multicolumn{1}{c|}{9.35}      &      --         &             --          &           --           \\
\multicolumn{2}{c|}{\textsc{SegBart}}        & 21.49         &          8.29                                 & \multicolumn{1}{c|}{19.52}      &       16.36 	
        &           5.14            &       14.96          \\    \multicolumn{2}{c|}{\textsc{SegUnilm}}                  & \bf{22.17}           & \bf{8.86}           & \multicolumn{1}{c|}{\bf{20.17} }     &         \bf{17.59}       &       \bf{6.20}                & \bf{15.90}                     \\
\thickhline

\end{tabular}
\caption{ROUGE F1 results on \textsc{SegNews} test set. R1 and R2 are shorthands for ROUGE scores of unigram and bigram overlap;  RL is the ROUGE score of longest common subsequence.  In pipeline approach, we try combinations of different segmentators and generators. Due to their failure on segmentation, non-pretraind models have very low  ROUGE scores with predicted segments,  and we do not compare them in the table.}
\end{table*}

\subsection{Joint Approach}
Instead of relying on a pipeline framework, we  can also tackle the segmentation-based summarization task with a single encoder-decoder neural model. This brings two main advantages. First, the encoders for segmentation and generation can be shared, benefiting both tasks as a multi-task learner.
Second, we can decode all summaries in an auto-regressive manner. In this way, when the decoder generates the $l$-th heading, it will be exposed to the 1st to $(l-1)$-th generated headings. This is considerately helpful since in a news article, many headings are highly related and coherent in their content. 


We use Transformer~\cite{vaswani2017attention} as base model for the encoder and decoder. Formally, the encoder maps a sequence of tokens in the source document
$\bm{x} = [x_1, ..., x_n]$ into a sequence of continuous representations
$\bm{t} = [t_1, ..., t_n]$.
Then a  segmentation classification layer is applied over output vectors of paragraph boundaries to identify correct segments $B=[b_1, b_2, \cdots, b_{N-1}]$ for the input article.
The decoder then generates the tokens
of the target text $\bm{y} = (y_1, ..., y_m)$ auto-regressively based on the conditional probability:
$p(y_1, ..., y_m|x_1, ..., x_n, B)$.
As the decoder produces summaries for all sections in one pass, we add a special symbol \texttt{[Y\_SEP]} between summaries from neighboring sections to indicate their boundaries.

However, in this vanilla sequence-to-sequence model, during inference, the decoder is not aware of the segmentation results and can only implicitly use this information when decoding the summaries. Thus, to better jointly learn segmentation and generation tasks, we propose \textsc{SegTrans}  model, which is equipped with Segmentation-aware Attention mechanism.  


\textbf{Segmentation-aware attention} The multi-head decoder-to-encoder attention in a Transformer decoder defines that for a head $z \in \{1,\cdots,n_{head}\}$ at each layer, the model calculates attention probabilities $a^z_{ij} $ against each source token $x_j$ when generating the $i$-th token $y_i$. 
    \begin{gather}
    q^z_i = W^z_qY_i,\\
    k^z_j = W^z_kX_j,\\
    a^z_{ij} =\frac{exp({q^z_i}^Tk^z_{j})}{\sum_{o=1}^n{exp({q^z_i}^Tk^z_{o})}},
    \end{gather}
    \vspace{1ex}
    where $Y_i, X_j\in \mathbb{R}^{d}$ are the layer's input vectors corresponding to the token $y_i$ and $x_j$, respectively.
    $W^z_q, W^z_k\in
    \mathbb{R}^{d_{head}*d}$ are learnable weights. $n$ is the number of tokens in source input.

However, in segmentation-based summarization, when generating the heading for the $i$-th section, the decoder should focus more on the input tokens belonging to that section. Thus, we propose the segmentation-aware attention as follows.

We select a subset $\hat{z}$ of decoder heads  to apply a segmentation mask to enforce that these heads only attend to the corresponding section. For a head in $\hat{z}$, Eq. 3 is modified to:
    \begin{gather}
    a^z_{ij} =\frac{exp({q^z_i}^Tk^z_{j})seg(y_i,x_j)}{\sum_{o=1}^n{exp({q^z_i}^Tk^z_{o})seg(y_i,x_j)}}
    \end{gather}
    where $seg(y_i,x_j)$ is a indicator function. It equals 1 if and only if $y_i$ and $x_j$ both belong to the same section, and 0 otherwise. In this manner, parts of the heads in multi-head attention are able to dynamically capture segmentation information, while the other heads still model global features of the entire input article.
    
We illustrate a detailed example of our framework with segmentation-aware attention in Figure~\ref{fig2}.
We first encode the source text, and apply a segmentation classification layer over output vectors of paragraph boundaries. For this example input, the model classifies the first and the third paragraph boundaries to be segmentation points. Then the decoder will apply a segmentation-aware multi-head attention over the source outputs. It  generates the summary for the first identified section with parts of the attention heads over only the first and the second paragraphs. After generating the first heading ending symbol  \texttt{[Y\_SEP]}, the decoder changes the segmentation-aware attention to the third paragraph for generating the summary for the second section.

The final loss $\mathcal{L}$ for training \textsc{SegTrans} is the summation of the segmentation loss (binary classification loss) $\mathcal{L}_{\text{seg}}$ and generation loss (negative likelihood loss)  $\mathcal{L}_{\text{gen}}$:
\begin{equation}
    \mathcal{L} =\mathcal{L}_{\text{seg}} + \mathcal{L}_{\text{gen}}
\end{equation}

One advantage of our framework is that it is a non-invasive adaptation of the Transformer model, i.e. it does not alter the inner structure of Transformers.  This is important since this adaptation can be applied to many popular pretrained language generation models (e.g. \textsc{Mass}, \textsc{Bart}, GPT and \textsc{UniLM}), offering our framework a high degree of flexibility and better performance. In this paper, we also augment pre-trained   \textsc{UniLM} model and \textsc{Bart} model with this mechanism and propose \textsc{SegUniLM}  and \textsc{SegBart} to further boost their performance.



\section{Experiments}
In this section, we conduct experiments on \textsc{SegNews} dataset by comparing our proposed model with several strong baselines.

\subsection{Experimental Settings}
In pre-processing, all
the words in news articles and headings are transformed to lower case and tokenized with wordpiece tokenizer from \textsc{BERT}~\cite{devlin2018bert}.
In data splitting, we guarantee the headings of articles in the test set have low bigram overlap with articles in the training set.
We obtain a splitting of 21,748 articles in  training set, 2,688 in  validation set and  2,444 in test set.

We experiment under both non-pretrained and pretrained settings.
In non-pretrained setting, we use a 6-layer Transformer encoder-decoder model  (\textsc{SegTrans})  with  512  hidden size and 2,048 feed-forward filter size.
In pretrained setting, we propose \textsc{SegUniLM} and \textsc{SegBart}  which adopts the base version of \textsc{UniLM}v2~\cite{bao2020UniLMv2}  and the large version of \textsc{Bart}~\cite{lewis2019Bart} as the
pretrained model.  \textsc{UniLM}v2 is a Transformer-based neural
network with 12 Transformer layers and 12
attention heads. It is pretrained as a pseudo-masked language model on
a large corpus.  
\textsc{Bart} is a Transformer-based neural encode-decoder model with 12 layers and 16 attention heads, pretrained via a denoising auto-encoder loss.
Label smoothing is used with smoothing factor~0.1.  
For segmentation-aware attention, we choose the best $c$ (number of segmentation-aware heads) by experiments on the validation set, and $c=9$ for \textsc{SegUniLM} and  $c=13$ for \textsc{SegBart} provide the best performance.

During all decoding we use beam search (size~5), and tune $\alpha$~for
the length penalty~\citep{wu2016google} between 0.6 and 1 on the
validation set.
To guarantee the number of generated headings can match the number of predicted source segments, we take a trick of only generating the end-of-generation token (\texttt{EOS}) when these two numbers match.

We compare the proposed joint models with two sets of strong baselines. The first set of baselines are vanilla sequence-to-sequence models. These models take raw  article as input and output the concatenated headings. The second set are pipeline models. As described, these systems first use a segmentor to divide the article into several sections, and then apply a generator to produce summary for each section.

In segmentation-based summarization, summarization systems require segmentation results. We set two settings of segmentation. 
For the first setting, we  provide golden segments to the models to evaluate their performance of generating the summaries when given the correct segments. 
For the second setting, we require the models to first segment the article and then generate summaries for the predicted segments.


\subsection{Evaluation Metrics}
Evaluation metrics for summarization performance are ROUGE~\cite{lin:2004:ACLsummarization} F1 scores of the generated headings against the gold headings.
We report unigram and bigram overlap (ROUGE-1 and ROUGE-2) as a means of assessing informativeness and the longest common subsequence (ROUGE-L) as a means of assessing fluency.

We use standard metrics \textit{Pk}~\cite{beeferman1999statistical} and \textit{WinDiff}~\cite{pevzner2002critique} to evaluate segmentation results. Lower scores of these two metrics indicate that the predicted segmentation is closer to the ground truth. A \textsc{Even} baseline is included for comparison where it segments the whole article evenly.


\begin{table}
\renewcommand\arraystretch{1}
\centering
\begin{tabular}{l|ccc}
\thickhline
Models                                                                     & R1 & R2 & RL \\ \hline
\textsc{SegUnilm}                                                                   & 22.17&8.86&20.17   \\\hline
\quad (c=12)                                                &  22.14&	8.81&	20.09
       \\
\quad (c=8)                                                         &  22.13&	8.84&	20.10
    \\
\quad (c=4)                                                             & 21.39&	7.99&	19.23
     \\
\quad (c=0)                                                            &  19.85 &	7.74 &	17.62
  \\
\quad (w/o seg loss) &  22.06&	8.66&	20.02
  \\
 \thickhline
\end{tabular}
\caption{ Ablation study results on \textsc{SegNews}. We compare multiple variants of \textsc{SegUniLM}. $c$ indicates the number of decoder heads modified into segmentation-aware attention. Be default, \textsc{SegUniLM} uses $c=9$ to achieve the best performance. We also present a \textsc{SegUniLM} model  without (w/o)  segmentation classification loss, and it is trained solely by generation loss. }\label{tab:ablation}
\end{table}

\begin{table}[]
\renewcommand\arraystretch{1}
\centering
\begin{tabular}{l|cc}
\thickhline
Model       & WD & PK \\ \hline
\textsc{Even}& 0.469&0.450\\
Transformer &      0.563      & 0.462 \\
\textsc{Bart}        &      0.484     &  0.411  \\
\textsc{Unilm}       &    0.479      &  0.391  \\
\textsc{SegBart}     &   0.471      & 0.405  \\
\textsc{SegUnilm}    &   0.462       &  0.380  \\
\thickhline
\end{tabular}
\caption{ Experimental results on document segmentation task. WD indicates \textit{WinDiff} metric.}\label{tab:segment}
\end{table}

\subsection{Results}

Table 2 describes our summarization results on the \textsc{SegNews} dataset. 
The first vertical block includes the results of vanilla sequence-to-sequence models. 
\textsc{Trans} is the non-pretrained Transformer encoder-decoder model.
\textsc{Unilm} and \textsc{Bart} are two pretrained baseline models.
The second vertical block contains the results of pipeline models. We present the combinations of different segmentation models and generation models.
For segmentor, we experiment non-pretrained Transformer model and pretrained \textsc{Bart} and \textsc{UniLM} models. For generator, we also include \textsc{TopicRank}, which is a classical extractive summarization method.

The last vertical block
includes the results of our joint models: \textsc{SegTrans},  \textsc{SegBart}  and \textsc{SegUniLM}. They  respectively rely on non-pretrained Transformer and pretrained \textsc{Bart} and \textsc{UniLM}   as backbone models. Segmentation-aware attention mechanism is used to augment these jointly trained systems.

We can see vanilla sequence-to-sequence models with no segmentation information input perform poorly on this task.
End-to-end \textsc{SegUniLM} model achieves the best performance among all systems. 
\textsc{SegUniLM} outperforms the best pipeline system under both settings when gold segments or predicted segments are provided.
This indicates \textsc{SegUniLM} has better overall performance and will be more useful when applied as practical applications.
It also  shows higher summarization results than vanilla \textsc{UniLM} model, confirming the effectiveness of segmentation-aware attention mechanism.
 \textsc{SegBart} and  \textsc{SegTrans} also show similar superiority over their pipeline versions.
Examples of system output are shown in Table \ref{table:examples}.

\begin{table*}
  \renewcommand{\arraystretch}{1.15}
  
  \small
\begin{center}
  \begin{tabular}{@{}l@{~~}p{13.8cm}@{}}\thickhline
      \cline{1-2}
    \multicolumn{2}{c}{{Title: One JFK conspiracy theory that could be true}} \\ \hline
      \textsc{Gold} &1. LBJ had it done; 2. The military industrial complex did it; 3. The mob did it; 4. Oswald acted alone as part of an unknown conspiracy; 5. The CIA did it
\\
      Pipeline \textsc{UniLM}         &   those kennedys will never embarrass me again; did kennedy want to withdraw us troops from vietnam ?; 3. different mobs; other conspirators ?; would america be ok with that ?  \\
      \textsc{SegBart}       &they thought he was a crook; he was going to pull american troops out of vietnam; the mob did this; there were others, but who were they?; the russians ordered the killing\\
            \textsc{SegUniLM}       &1. those kennedys will never embarrass me again; 2. he said he'd pull troops out of vietnam; 3. mob members claim they were witnesses to the alleged shootings; 4. there were more people who knew where oswald was; 5. the cia didn t release any of the good stuff
  \\ \thickhline
      \multicolumn{2}{c}{{Title: What is the diversity visa lottery?}}                                                                                                  \\ \hline
     \textsc{Gold}     & What is it?; How does it work?; How did it get started?; Has Congress tried to change the program?; What is Trump proposing?\\
    Pipeline \textsc{UniLM} &what is a green card?; how the program works; history of the visa program; schumer helped replace the program; create a point system ...\\
      \textsc{SegBart}  & what is the diversity visa program?; how does the program work?; who created the program?; who has sought reform?; what are the next steps?\\
            \textsc{SegUniLM}  & what is the diversity visa program?; what are the requirements for the visas?; how did it start?; was the "diversity visa" created by the gang of eight?; is there any debate over reform?\\
    \thickhline
      \multicolumn{2}{c}{{Title: This man is tasked with finding out who failed Larry Nassar's victims}}
    \\ \hline
     \textsc{Gold}  & Seeking justice; A very youthful 68-year-old; A model independent prosecutor\\

      Pipeline \textsc{UniLM}                         & searching for truth; he couldn't stay retired; he didn't have an agenda
      \\ 
     \textsc{SegBart}          & searching for the truth; working with juveniles; no stone unturned\\ 
          \textsc{SegUniLM}          & searching for the truth; he's has to do something; he doesn't have an agenda\\ \thickhline

  \end{tabular}
\end{center}
\vspace*{-2ex}
\caption{\textsc{Gold} reference summaries and automatic summaries
    produced by  pipeline \textsc{UniLM}, \textsc{SegBart} and  \textsc{SegUniLM} on the \textsc{SegNews}  datasets. Semicolons indicate the boundaries of headings.  }\label{table:examples}
\end{table*}

\begin{table}
\renewcommand\arraystretch{1}

\begin{center}

\begin{tabular}{l|cc}

\thickhline
Model          & Quality & Fluency \\ \hline
Pipeline \textsc{UniLM} &     1.93    &   2.62     \\
\textsc{SegUniLM}       &   2.17      &    2.59    \\
Gold           &    2.44     &      2.79 \\
\thickhline

\end{tabular}
\end{center}
\caption{Human evaluation results based on summary quality and fluency.}
\end{table}

    Table~\ref{tab:ablation} summarizes   ablation studies
    aiming to assess the contribution of individual 
    components of  \textsc{SegUniLM}.
    We first modify \textsc{SegUniLM}  by varying $c$, the number of heads of segmentation-aware attention. We can see the best results of ROUGE are achieved when $c=9$. With more or less heads modified as segmentation-aware attention heads, the summarization performance show a clear trend of decreasing.
    Also, as shown in the last column, when segmentation layer and segmentation loss are removed, we observe a sharp decrease on ROUGE scores. The results prove that both segmentation-aware attention and joint training provide improvement to the summarization results.
    
    Table~\ref{tab:segment} describes  the results on news segmentation task.
    \textsc{SegUniLM} achieves the lowest WD and PK scores, revealing its ability to identify the structure of a news article.
    Compared with \textsc{Unilm} model without the segmentation-aware attention, \textsc{SegUniLM} shows clear superiority on both metrics. The same trend is also observed in \textsc{Bart} related models.

\subsection{Human Evaluation}
    In addition to automatic evaluation, we also assess system
    performance by eliciting human judgments on 20 randomly selected
    test instances.
    The evaluation study assess the overall quality and fluency of the
    summaries by asking participants to rate them. 
    We present the news article to evaluators along with system generated heading-style summaries, and we ask evaluators to read the complete article, and give scores based on summary quality  and fluency respectively. 
    Participants can have three scores (1-low quality/fluency, 2-median quality/fluency, 3-high quality/fluency).

    Gold summaries, outputs from pipeline \textsc{UniLM} and  \textsc{SegUniLM} models are compared in evaluation.
    We invite three evaluators with linguist background to conduct the human evaluation. The averaged results are shown in Table 4. 
     Overall, we observe pipeline \textsc{UniLM} and  \textsc{SegUniLM} perform similarly on fluency, but   \textsc{SegUniLM}  shows its superiority on summary quality.
     Gold summaries are marginally better than automatic generated summaries.

\section{Conclusion}

In this work, we proposed a new task, segmentation-based news summarization.
It aims to segment a news article into multiple sections and generate the corresponding summary to each section.
This new task provides a novel alternative to digesting a news article.
We built a new
benchmark dataset \textsc{SegNews} to study and evaluate the task.
Furthermore, we designed a segmentation-aware attention mechanism, which allows neural decoder to capture segmentation information in the source texts.
The new attention mechanism is a non-invasive adaption of Transformer models and can be integrated with many pretrained language generation models.
We jointly train the model for generating summaries and recognizing news segments.
Experimental results on  \textsc{SegNews} demonstrate that our framework produces
    better segmentation-based summaries than
    competitive systems.

\newpage
\bibliography{aaai22}

\end{document}